\pdfoutput=1

\documentclass[11pt]{article}

\usepackage[final]{acl}

\usepackage{times}
\usepackage{latexsym}
\usepackage{adjustbox}
\usepackage{multirow}
\usepackage{multicol}
\usepackage{amsmath}
\usepackage[ruled,vlined]{algorithm2e}
\usepackage{colortbl}
\usepackage{float}
\definecolor{lightgray}{gray}{0.9}
\definecolor{headergray}{gray}{0.8}
\definecolor{darkgray}{gray}{0.6}

\usepackage[T1]{fontenc}

\usepackage[utf8]{inputenc}

\usepackage{microtype}

\usepackage{inconsolata}

\usepackage{graphicx}

\title{Verify-in-the-Graph: Entity Disambiguation Enhancement for Complex Claim Verification with Interactive Graph Representation}

\author{
  Hoang Pham, 
  Thanh-Do Nguyen, and
  Khac-Hoai Nam Bui\thanks{Corresponding author} \\
  Viettel Artificial Intelligence and Data Services Center, \\
  Viettel Group, Vietnam \\
  \{hoangpv4, dont15, nambkh\}@viettel.com.vn
}

\begin{document}
\maketitle

\begin{abstract}
Claim verification is a long-standing and challenging task that demands not only high accuracy but also explainability of the verification process. This task becomes an emerging research issue in the era of large language models (LLMs) since real-world claims are often complex, featuring intricate semantic structures or obfuscated entities. Traditional approaches typically address this by decomposing claims into sub-claims and querying a knowledge base to resolve hidden or ambiguous entities. However, the absence of effective disambiguation strategies for these entities can compromise the entire verification process. To address these challenges, we propose Verify-in-the-Graph (VeGraph), a novel framework leveraging the reasoning and comprehension abilities of LLM agents. VeGraph operates in three phases: (1) Graph Representation - an input claim is decomposed into structured triplets, forming a graph-based representation that integrates both structured and unstructured information; (2) Entity Disambiguation -VeGraph iteratively interacts with the knowledge base to resolve ambiguous entities within the graph for deeper sub-claim verification; and (3) Verification - remaining triplets are verified to complete the fact-checking process. Experiments using Meta-Llama-3-70B (instruct version) show that VeGraph achieves competitive performance compared to baselines on two benchmarks HoVer and FEVEROUS, effectively addressing claim verification challenges. Our source code and data are available for further exploitation\footnote{https://github.com/HoangHoang1408/VeGraph}.
\end{abstract}

\section{Introduction}
\begin{figure}[!h]
  \centering
  \includegraphics[width=\columnwidth]{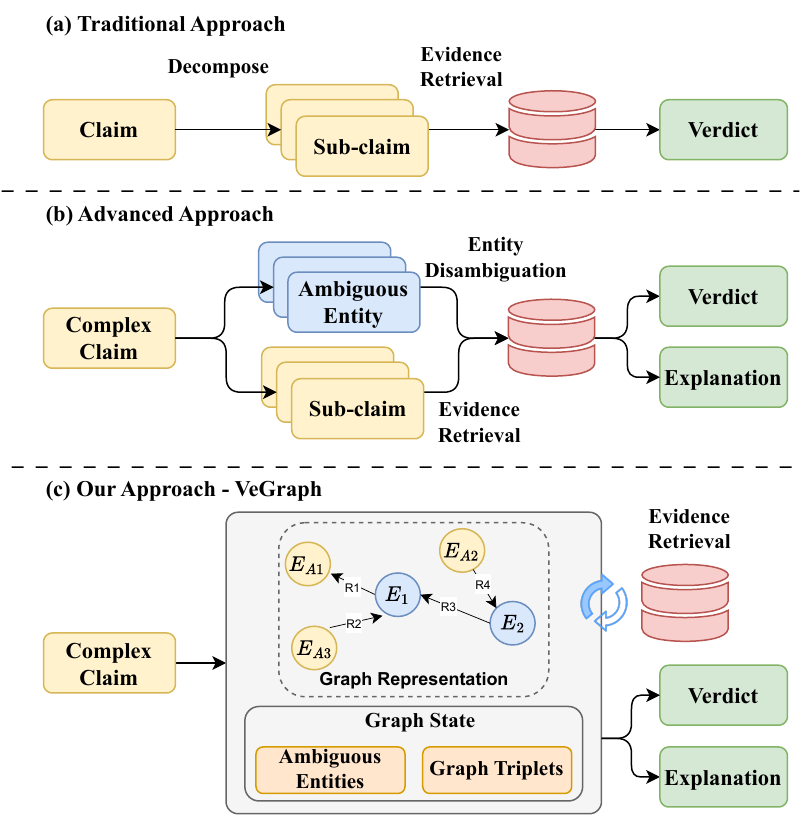}
  \caption{Conceptual analysis of previous works and VeGraph: a) Traditional approaches use IR to retrieve evidence and then verify sub-claims; b) Advanced approaches use IR to resolve ambiguous entities and then verify sub-claims; c) Our approach represents claims with graph triplets, then iteratively interacts with IR for entity disambiguation and sub-claims verification.}
  \label{fig:comparison}
\end{figure}

In the era of rapidly advancing large language models (LLMs), the widespread dissemination of misinformation, combined with the increasing presence of AI-generated content, has made it significantly harder for individuals to assess the reliability of information. Consequently, claim verification, leveraging advanced Natural Language Processing (NLP) techniques to automatically determine the veracity of claims, has emerged as a critical research topic \cite{fact_check_survey, abs-2408-14317}.

Traditional approaches typically begin by decomposing a given claim (e.g., at the sentence or passage level) into sub-claims, often using methods such as chain-of-thought (CoT) prompting \cite{Wei0SBIXCLZ22}. Subsequently, each sub-claim is evaluated by prompting an LLM, incorporating knowledge sources (e.g., information retrieval systems) to determine the truthfulness of the overall claim \cite{Krishna0022, ZhangG23}, as shown in Figure \ref{fig:comparison}(a). Multi-step reasoning in LLMs is the process of addressing complex tasks by breaking them into sequential inference steps, where each step builds on the previous one, enabling the model to integrate intermediate results and draw conclusions. Recently, more advanced methods have enhanced claim verification task by incorporating multi-step reasoning to resolve ambiguous entities before verifying sub-claims \cite{WangS23a, program_fc, ZhaoWWCZW24}, as illustrated in Figure \ref{fig:comparison}(b). These improvements have made such methods more promising for explainable and interpretable claim verification systems.

However, despite the advancements achieved by multi-step reasoning mechanisms, several critical challenges persist: i) \textit{Ambiguous Entity Interactions:} Ambiguities in entity relationships remain a significant hurdle for fact verification systems \cite{sedova2024knowknowanalyzingselfconsistency}. This challenge is amplified in multi-step reasoning, where entity disambiguation must span the entire verification process. Unlike previous approaches that employ external tools for resolving ambiguities in individual sub-claims, effective resolution here requires seamless integration throughout the reasoning pipeline; ii) \textit{Limitations of LLM-Based Multi-Step Reasoning Agents:} Many existing approaches rely on static, single-plan veracity prediction \cite{program_fc, WangS23a}. If a failure occurs at any intermediate step, the entire reasoning process may collapse, thereby underutilizing the adaptive potential of LLM-based agents to recover and refine reasoning paths dynamically.

In response to these challenges, this study introduces an agent-based framework, named Verify-in-the-Graph (VeGraph), for automatic fact verification. Our approach, illustrated in Figure \ref{fig:comparison}(c), consists of three interconnected stages: an LLM agent first constructs a graph-based representation by decomposing the input claim into sub-claim triplets. The agent then interacts with a knowledge base to resolve ambiguous entities in triplets, iteratively updating the graph state. Finally, the agent verifies triplets, completing the process. Overall, the primary contributions of this work are as follows:

\textbf{(1)} We propose a novel multi-step reasoning approach for claim verification using an LLM agent framework with interactive graph representation (VeGraph). To the best of our knowledge, this is the first study to leverage multi-step reasoning in conjunction with an interactive entity disambiguation process to enhance claim verification performance.

\textbf{(2)} The proposed method, by integrating interactive graph representations with LLM agent frameworks, enhances explainability and interpretability by exploiting both structured and unstructured information — the key elements for advancing multi-step reasoning tasks.

\textbf{(3)} We evaluate and show the effectiveness of our approach on two widely recognized benchmark datasets in this research field: HoVer \cite{hover_dataset} and FEVEROUS \cite{feverous_dataset}. 

\section{Related Work}
\begin{figure*}[!h]
  \centering
  \includegraphics[width=\textwidth]{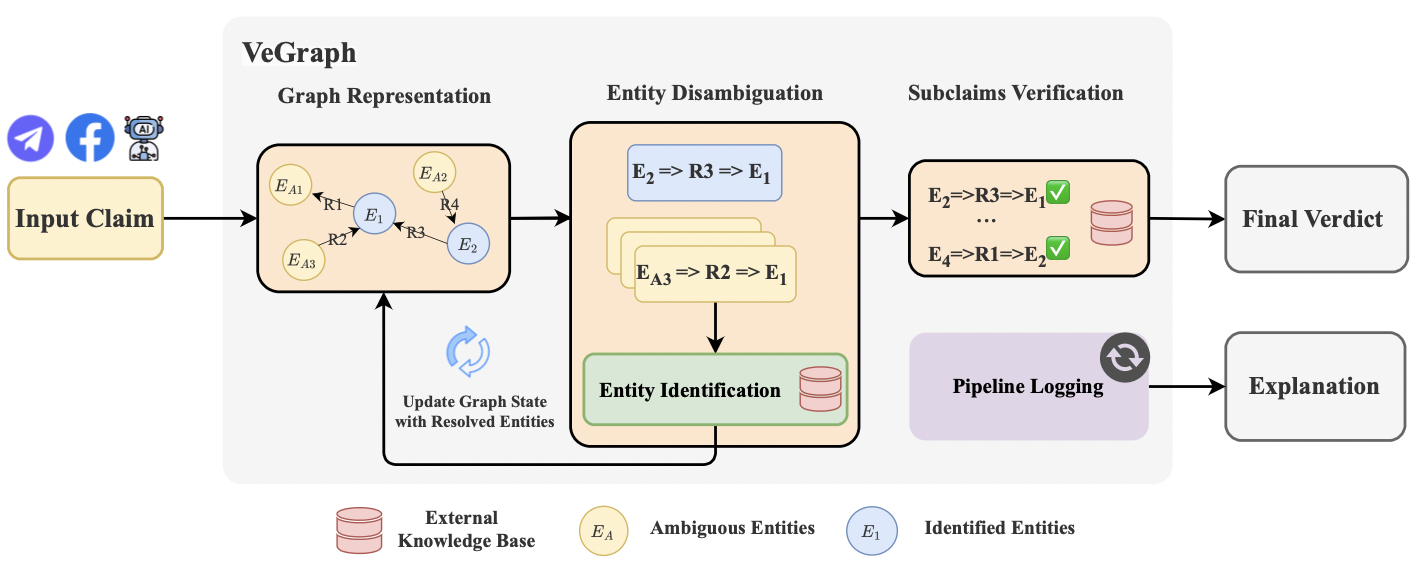}
  \caption{Three key components of VeGraph: (i) \textbf{Graph Representation}, which decomposes the complex input claim into graph triplets; (ii) \textbf{Entity Disambiguation}, ambiguous entities are resolved through iterative interactions with the knowledge base (KB); and (iii) \textbf{Sub-claim Verification}, which evaluates each triplet by delegating the verification process to the sub-claim verification function. The logging module records the whole process.}
  \label{fig:main_pipeline}
\end{figure*}

Claim verification is a long-standing and challenging task that seeks to determine the veracity of a claim by retrieving relevant documents, selecting the most salient evidence, and making a veracity prediction. In the era of large language models (LLMs), LLM-based claim verification has evolved to generate subclaims from input claims using the chain-of-thought (CoT) approach, and to retrieve evidence by augmenting the LLM with external knowledge sources for verification \cite{fact_check_survey}. ProgramFC \cite{program_fc} improves this process by leveraging in-context learning along with the CoT method, decomposing the original claim into program-like functions to guide the verification steps. Similarly, FOLK \cite{WangS23a} translates the claim into First-Order-Logic (FOL) clauses, where each predicate corresponds to a subclaim that requires verification. FOLK then performs FOL-guided reasoning over a set of knowledge-grounded question-answer pairs to predict veracity and generate explanations, justifying its decision-making process. Furthermore, PACAR \cite{ZhaoWWCZW24} leverages LLM Agent concept, which incorporates self-reflection technique and global planning to enhance performance. 

Despite the advancement of these methods, which exploit LLM reasoning capabilities to interact with external knowledge bases, they are limited to a single interaction with the knowledge base for an ambiguous entity. If the knowledge base fails to identify the requested entity in the query, the entire verification process may collapse. In light of these limitations, our proposed method similarly leverages LLM reasoning in conjunction with external knowledge retrieval systems. However, we extend this by incorporating agent-based LLM, enabling iterative interactions with the knowledge base to resolve ambiguous entities and execute multi-step reasoning for more robust and in-depth claim verification.
 
\section{Methodology}
The main objective of this study is to predict the veracity of a complex input claim $C$ through automated reasoning using an interpretable LLM Agent, incorporating both structured and unstructured information through graph representation. Figure \ref{fig:main_pipeline} shows the architecture of our proposed framework. Specifically, VeGraph consists of three stages: (i) the agent represents the claim $C$ with graph triplets, each corresponding to a sub-claim; (ii) the agent interacts with an external knowledge base to resolve ambiguous entities; and (iii) once all ambiguities are addressed, the agent verifies sub-claims corresponding to the remaining triplets. The veracity of the input claim is determined by the veracity of all graph triplets, if all the graph triplets are verified with the information in the knowledge base then the claim $C$ is $Supported$, if one of the triplets cannot be verified then the claim $C$ is $Refuted$. During processing through stages, the logging module records the activities of the agent for explainability.

\subsection{Graph Representation}
\begin{figure}[!h]
  \centering
  \includegraphics[width=0.93\columnwidth]{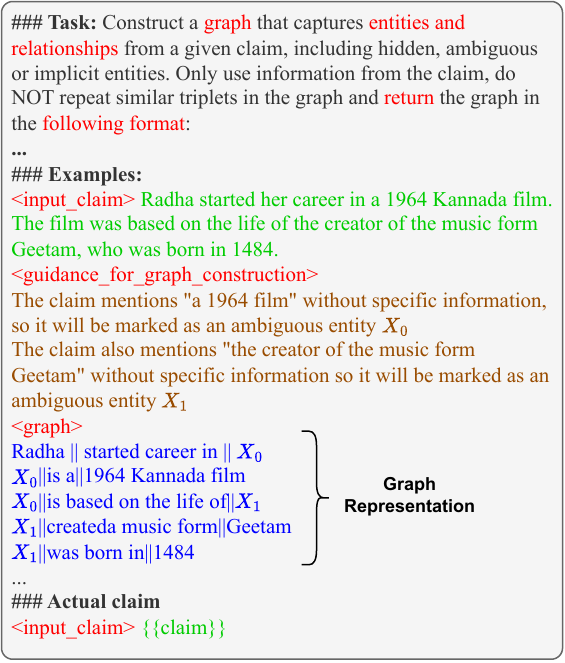}
  \caption{Prompt to make LLM construct the Graph Representation}
  \label{fig:construct_graph}
\end{figure}
Input claims often contain complex sentence structures that challenge LLMs to grasp their semantic meaning. To address this, we transform each claim into a graph representation composed of triplets, with each triplet capturing a subclaim within the original claim (illustrated in Figure \ref{fig:construct_graph}). This semantic graph construction is grounded in techniques from the field of Information Extraction, utilizing a joint approach for entity and relation extraction \cite{join_extract_1, join_extract_2} in an end-to-end fashion. Entities (nodes) are defined as spans of text that represent objects, events, or concepts mentioned in the claim. Unlike traditional Named Entity Recognition (NER) systems, which rely on fixed categories, this approach accommodates a more diverse set of entity types. For relation extraction (edges), we apply methods from Open Information Extraction (OpenIE) \cite{extract_relation} leveraging LLMs' semantic comprehension. Instead of restricting relations to predefined categories (e.g., OWNERSHIP, LOCATED), this method extracts relations expressed in natural language, capturing detailed document-level interactions. For instance, in a semantic graph, a relation like “is based on the life of” (in Figure \ref{fig:construct_graph}) accurately represents the relationship between two entities within the claim.

Formally, in \textbf{VeGraph}, the graph construction process leverages in-context learning \cite{Wei0SBIXCLZ22} to prompt the LLM to generate graph $G = \{T_1, T_2, ..., T_N\}$ consisting of N triplets, each triplet $T_i = (E_{1i}, R_i, E_{2i})$ corresponds to a subclaim extracted from the original claim $C$. Here, $E_{1i}$ and $E_{2i}$ denote the head and tail entities, respectively, while $R_i$ captures the semantic relation between them. Complex claims often contain implicit or ambiguous entities that need to be resolved to facilitate claim verification. For example, in the claim shown in Figure \ref{fig:construct_graph}, the entity “a 1964 Kannada film” is not explicitly named, necessitating a disambiguation process. To address this, we categorize entities into two types: explicitly stated entities are marked as standard entity nodes, while ambiguous entities are tagged as $X_i$ to signal the need for further clarification. This disambiguation process of these entities, detailed in Section \ref{sec:entity_disambiguation}, ensures a comprehensive representation of claim semantics. With this graph-based representation, the LLM can more effectively capture the semantic intricacies of the claim, thereby enhancing its reasoning capabilities and supporting improved performance of claim verification. (Refer to Figure \ref{fig:few_shot_construct_graph} in Appendix for the detailed prompt)

\subsection{Knowledge Base Interaction Functions}
\label{interaction_function}
To facilitate interaction with the knowledge base in the open-book setting, we implement two core functions: \textit{Entity Identification} and \textit{Claim Verification}. Both functions utilize Information Retrieval techniques to retrieve relevant documents enabling context-aware decision-making. During execution, all the retrieved documents are recorded for thoroughness and explainability.
\\
\textbf{Entity Identification.} This function acts as a question-answering module that extracts a specific entity. Formally, for a given question $Q$, a set of top-$k$ relevant documents $D$ are retrieved from the knowledge base using an information retrieval system. The question $Q$ and the retrieved documents $D$ are processed jointly by the LLM to identify the target entity requested in the question. This allows the system to leverage external knowledge to resolve ambiguities and produce informed answers. (Refer to Figure \ref{fig:qa_with_docs} in Appendix for the prompt)
\\
\textbf{Sub-claim Verification.} The \textit{Sub-claim Verification} function is designed to assess the truthfulness of a given claim $C$. Upon receiving a claim as input, the system retrieves a set of top-$k$ documents $D$ relevant to $C$ from the knowledge base. These documents are then processed alongside the claim by the LLM, which determines whether the information supports or refutes the claim. The output is a binary decision—either \textit{True} or \textit{False}—that indicates the veracity of the sub-claim (Refer to Figure \ref{fig:fact_check_with_docs} in Appendix for the detailed prompt).

\subsection{Entity Disambiguation Process}
\label{sec:entity_disambiguation}
\begin{figure*}[!h]
  \centering
  \includegraphics[width=\textwidth]{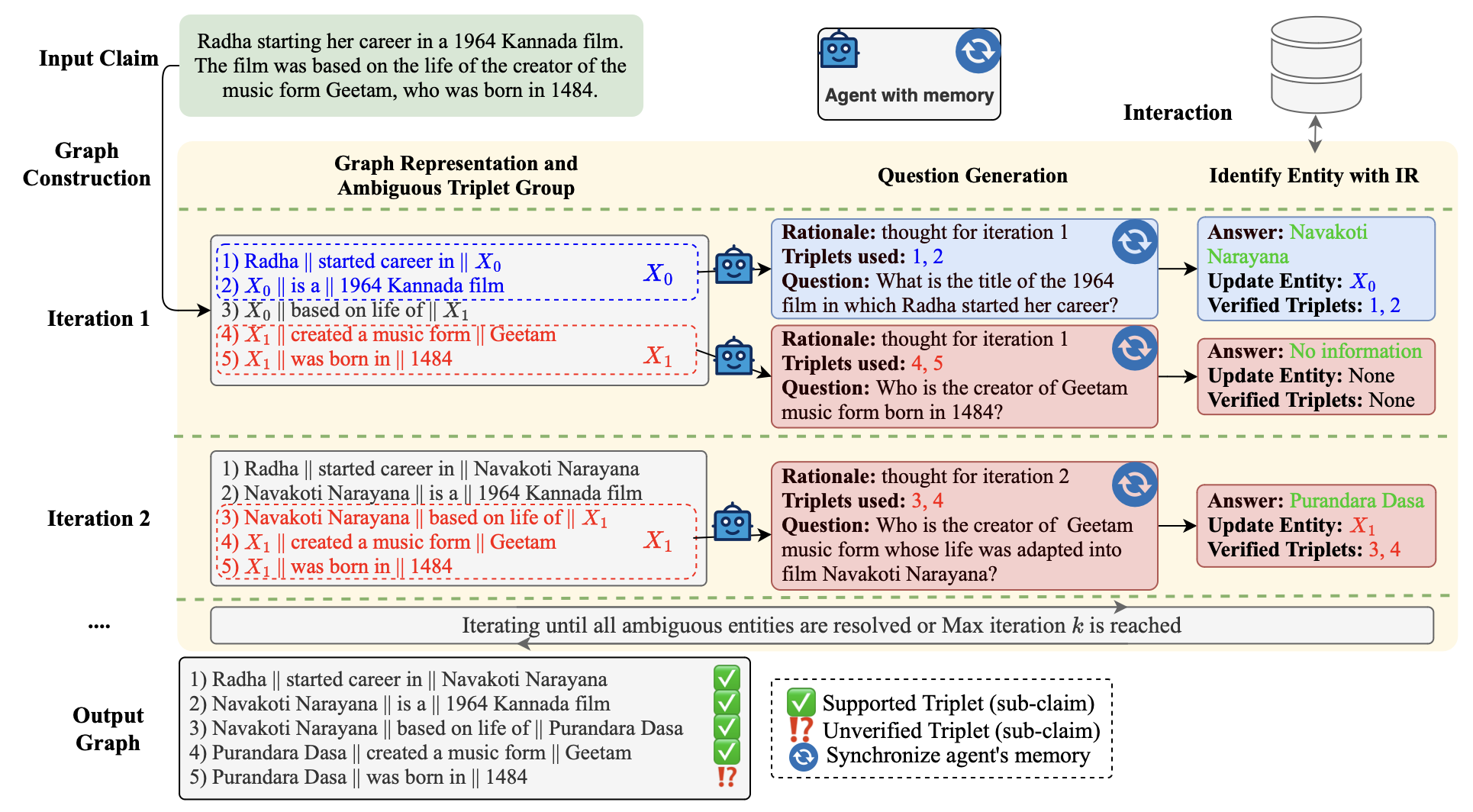}
  \caption{Illustration of the entity disambiguation process}
  \label{fig:disambiguation_process}
\end{figure*}

Following the transformation of the claim into a graph representation, the next step is identifying and resolving ambiguous entities. The disambiguation process is described in Algorithm \ref{algo:disambiguation} and illustrated step-by-step in Figure \ref{fig:disambiguation_process}.
\\
\textbf{Triplet Grouping.}
To effectively address entity ambiguities, we organize the extracted triplets from the graph $G$ into distinct groups based on shared ambiguous entities. Each group consists of triplets containing the same ambiguous entity. For instance, in Figure \ref{fig:disambiguation_process}, the triplets are grouped according to two ambiguous entities, $X_1$ and $X_2$. This method isolates each ambiguous entity along with relevant information, facilitating a more focused resolution.
\\
\textbf{Interaction with Knowledge Base.}
Once the triplets are grouped, the LLM interacts with each group to generate clarifying questions for the ambiguous entities. A major challenge arises when entity-related information in the knowledge base is often fragmented across multiple documents or sections, leading to that if we combine all the information or aspects related to an entity to find it from a specific partition of the knowledge base can be difficult. To address this, we adopt an iterative question refinement approach where the LLM uses the triplet information to narrow down ambiguities. Specifically, in each iteration, the LLM processes a group $g$ of triplets, producing the following outputs: i) a rationale $r$, which outlines the reasoning for selecting specific triplet information to construct the question; ii) a set of triplet identifiers $ids$, denoting the triplets used in formulating the question; and iii) a targeted question $q$, designed to clarify the ambiguous entity. The rationale $r$ guides the LLM in filtering relevant triplets ($ids$) for constructing a precise question $q$. This dynamic and self-controlled process enables the LLM to consider various aspects of the triplet group, ensuring comprehensive coverage of the information. The question $q$ is then processed by the function \textbf{\textit{Entity Identification}} to resolve the ambiguous entity.
\\
In addition, in the case when the question $q$ fails to resolve the ambiguity, this question along with its rationale are fed back into the LLM at the next iteration to generate a refined question $q'$ that incorporates alternative triplet aspects. As the process iterates, after each iteration, if an ambiguous entity $X$ in a group is clarified, the graph $G$ is updated accordingly by replacing $X$ with the actual entity founded. Other groups that have triplets related to $X$ benefit from this update, improving question refinement for those groups in subsequent iterations. For example, in Figure \ref{fig:disambiguation_process}, after the first iteration entity $X_0$ is identified as "Navakoti Nrayana", this information is then used to update other triplets (e.g. triplet with id 3). At the next iteration, this resolved entity adds more information related to the $X_1$ group. The iteration continues until either: i) all ambiguous entities are resolved; or ii) a maximum iteration limit $k$ is reached. The iterative refinement provides opportunities for the system to interact with the knowledge base and resolve the required ambiguous entity under a limited computing budget (Refer to Figures \ref{fig:generate_question} and \ref{fig:refine_question} for the prompts).
\\
\textbf{Verified Information and Outcome.}
When a question resolves an entity’s ambiguity, the corresponding triplets (with $ids$) are marked as containing verified information. The disambiguation process concludes when all ambiguous entities are resolved. If an entity remains ambiguous after $k$ iterations, the entire claim associated with that entity is classified as "REFUTES", indicating insufficient information for verification. Once all ambiguities are resolved, the disambiguation process outputs an updated graph with: i) \textbf{Verified triplets}: Triplets that contributed to the process of resolving ambiguities; and ii) \textbf{Remaining triplets}: Triplets that did not participate in the disambiguation process.

\subsection{Verification of Remaining Sub-claims}
After entity disambiguation, some triplets remain unverified, while others were not initially grouped for the disambiguation process. These remaining triplets require further verification. To achieve this, we employ a large language model (LLM) to generate full-text sub-claim questions based on the unverified triplets. For example, consider the triplet from Figure \ref{fig:disambiguation_process}: \textit{"Purandara Dasa || was born in || 1484"}. The LLM transforms this triplet into a full-text subclaim, such as \textit{"Purandara Dasa is the person who was born in 1484"}. This subclaim is then used in conjunction with the knowledge base for verification, facilitated by the \textbf{\textit{Subclaim Verification}} function. Once all remaining sub-claims are verified, the original claim $C$ is classified. If all sub-claims are supported, $C$ is categorized as $Supported$; otherwise, if any sub-claim is refuted, $C$ is categorized as $Refuted$.
\begin{algorithm}[h]
\small
\caption{Entity Disambiguation}
\label{algo:disambiguation}
\SetKwFunction{GroupTriplets}{GroupTriplets}
\SetKwFunction{AmbiguousEntities}{AmbiguousEntities}
\SetKwFunction{Main}{Main}
\SetKwFunction{GenQues}{GenQues}
\SetKwFunction{UpdateState}{UpdateState}
\SetKwFunction{Clarified}{Clarified}
\SetKwFunction{QA}{QA}
\SetKwFunction{Retrieve}{Retrieve}
\SetKwFunction{GenQuesAndResEntity}{GenQuesAndResEntity}
\SetKwFunction{Read}{Read}
\SetKwProg{Fn}{Function}{:}{}

\SetKwInOut{Input}{Input}
\SetKwInOut{Output}{Output}
\Input{Claim $C$, Input graph $G$, Max iteration $k$}
\Output{Clarified graph $G$, Verified triplets $VTriplets$}

\KwSty{Initialize:}\\
\hspace{3mm} Agent Attempt Logs: $logs = \emptyset$;\\
\hspace{3mm} Verified Triplets: $VTriplets = \emptyset$;\\
\Fn{\Main{$C, G, k$}}{
    \textit{// Logic of the disambiguation process} \\
    \For{$i = 1$ \KwTo $k$}{
        $groups =$ \GroupTriplets{$G$}\;
        \ForEach{$(ae, g)$ in $groups$}{
            \GenQuesAndResEntity{$ae, g$}
        }
    }
    \If{\Clarified{$G$}}{
        \textit{// check if all ambiguous entities is identified} \\
        \Return{"Successful"}\;        
    }
    \Return{"Failed"}\;
}
\Fn{\GenQuesAndResEntity{$ae, g$}}{
    \textit{// Agent try to generate question $q$ to identify the ambiguous entity $ae$ of the group $g$} \\
    $r, q, ids =$ \GenQues{$C, g, log[ae]$}\;
    $e =$ \QA{$q$}\;
    \If{$e \neq \texttt{None}$}{
        $VTriplets.\textbf{add}(ids)$\;
        \UpdateState{$G, ae, e$}\;
        \textit{// Update verified triplets and the state of the graph when identified a new entity} \\
    }
    \Else{
        $logs[ae].add((r, q))$\;
        \textit{// Log the rationale and the question when the agent failed}
    }
}
\Fn{\GroupTriplets{$G$}}{
    \textit{// Group triplets for ambiguous entities} \\
    $groups = \emptyset$\;
    $entities =$ \AmbiguousEntities{$G$}\;
    \ForEach{$ae$ in $entities$}{
        $group = \emptyset$\;
        \ForEach{$triplet$ in $G$}{
            \If{$ae \in triplet$}{
                $group.add(triplet)$\;
            }
        }
        $groups.add((ae, group))$\;
    }
    \Return{$groups$}\;
}
\end{algorithm}

\section{Experiments}
\begin{table*}[!h]
\centering
\begin{adjustbox}{width=\textwidth}
\begin{tabular}{l|c|ccc|ccc}
\hline\hline
\multirow{2}{*}{\textbf{Method}} &\multirow{2}{*}{\textbf{IR System}} &\multicolumn{3}{c|}{\textbf{HoVer}} & \multicolumn{3}{c}{\textbf{FEVEROUS}} \\

 & &\textbf{2hop} & \textbf{3hop} & \textbf{4hop} & \textbf{Multi-hop} & \textbf{Disambiguation} & \textbf{Numerical} \\
 \hline
 \rowcolor{lightgray}
\multicolumn{8}{c}{\textbf{Backbone LLM: GPT-3.5 Turbo (175B) or Codex; Different Experimental Setups}} \\   \hline 
\textbf{ProgramFC}* \cite{program_fc} & BM25 & 70.30 & 63.43 & 57.74 & - & - & - \\
\textbf{FOLK}* \cite{WangS23a} &  SERP API & 66.26 & 54.8 & 60.35 & 67.01 & - & 59.49 \\
\hline
\rowcolor{lightgray}
\multicolumn{8}{c}{\textbf{Backbone LLM: Meta-Llama-3-Instruct (70B); Same Experimental Setup}} \\   \hline 
\textbf{CoT-Decomposing} & Bi-Encoder &67.97 & 62.45 & 46.21 & 57.81 & 60.51 & 50.56 \\
\textbf{ProgramFC} (1 run) & Bi-Encoder&68.00 & 62.26 & 53.96 & 64.32 & 67.11 & 72.01 \\
\textbf{ProgramFC} (5 runs ensembled) & Bi-Encoder& \textbf{71.48} & 65.88 & 53.21 & \textbf{65.37} & 71.93 & 77.61 \\
\textbf{FOLK} & Bi-Encoder&67.74 & 58.49 & 53.47 & 60.96 & 61.00 & 47.44 \\
\hline
\textbf{VeGraph ($k = 5$)} & Bi-Encoder&\ 69.70 & \textbf{66.13}& \textbf{58.59} & 59.39 & \textbf{73.89} & \textbf{82.60} \\
\textbf{VeGraph ($k = 5$)} &  BM25& 69.22 & 63.10 & 56.68 & 53.29 & 72.46 & 82.06 \\
\hline\hline
\end{tabular}
\end{adjustbox}
\caption{Report results of Macro-F1 score on HoVer and FEVEROUS datasets. * are taken from respective papers. Both texts indicate the best score for the same experimental setup. }
\label{tab:main_results}
\end{table*}
\subsection{Datasets and Evaluation Metric}
\textbf{Dataset.} We conduct our experiments using an open-book setting, simulating a real-world scenario where the system has to interact with an external knowledge base to verify claims. We evaluate the proposed \textit{VeGraph} on two widely-used benchmark datasets for complex claim verification: HoVer and FEVEROUS. Both datasets contain intricate claims that require multi-hop reasoning and evidence gathering from various information sources. Due to the unavailability of public test sets, we rely on validation sets for evaluation. The HoVer dataset \cite{hover_dataset} is a multi-hop fact verification benchmark designed to validate claims using evidence across multiple sources, including 2-hop, 3-hop, and 4-hop paths. It is based on the introductory sections of the October 2017 Wikipedia dump. The multi-hop nature of HoVer challenges the system to retrieve and aggregate information from several interrelated documents. The FEVEROUS dataset \cite{feverous_dataset} addresses complex claim verification using both structured and unstructured data. Each claim is annotated with evidence derived from either sentences or table cells within Wikipedia articles of the December 2020 dump. For consistency with prior work \cite{feverous_dataset}, we evaluate FEVEROUS claims on three key partitions: Multi-hop Reasoning, Entity Disambiguation, and Numerical Reasoning. As our research focuses on textual fact-checking, we exclusively select claims that require sentence-based evidence, discarding those involving table cells or other structured data. To manage computational costs, specifically for the HoVer dataset, we sample 200 claims from each partition while ensuring balanced label distributions.
\\
\textbf{Metrics.} Following practices in the field, we use the Macro-F1 as the primary evaluation metric.

\subsection{Baselines}
For the comparison, we selected recent modern methods using LLM for multi-step reasoning veracity prediction, which are related to our work, as the baselines. Specifically, the baselines are sequentially described as follows:
\\
\textbf{CoT-Decomposing} CoT reasoning \cite{Wei0SBIXCLZ22} is a popular prompting approach that includes chains of inference steps produced by LLMs. Accordingly, for the claim verification task, the input claim is directly decomposed into subclaims using an LLM. These subclaims are then verified sequentially by prompting the LLM with facts grounded on external knowledge sources via the information retrieval systems. 
\\
\textbf{ProgramFC} \cite{program_fc} is one of the first claim verification models in the era of LLMs with the explainable capability for multi-step reasoning of veracity prediction. Specifically, the model decomposes complex claims into simpler sub-tasks and then solves the sub-tasks by using specialized functions with program-guided reasoning.
\\
\textbf{FOLK}  \cite{WangS23a} improve the explainable claim verification by introducing the first-order-logic (FOL) clause as the guided claim decomposition to make veracity predictions and generate explanations to justify step-by-step the verification decision-making process.

\subsection{Experimental Setups}
\textbf{Configurations:} Since the original baselines have different configurations including input data, information retrieval systems, and underlying LLM in their respective papers, therefore, we try to reproduce the baseline with the unified configuration, following their available source codes\footnote{https://github.com/teacherpeterpan/ProgramFC}\footnote{https://github.com/wang2226/FOLK}. To account for computational constraints, we limit the number of iterations $k$ in our proposed method, VeGraph, to 5. For a fair comparison, we also report the ensembled performance of ProgramFC over 5 runs, consistent with the original implementation \cite{program_fc}.
\\
\textbf{Backbone LLM and Prompting Strategy:} In our experiments, we employ Meta-Llama-3-70B-Instruct\footnote{\url{https://huggingface.co/meta-llama/Meta-Llama-3-70B-Instruct}} as the underlying LLM. To construct graph representations, we leverage in-context learning by providing the model with human-crafted examples to guide the LLM to perform the required tasks. For other tasks, we use zero-shot prompting leveraging existing LLM reasoning capability.
\\
\textbf{Retrieval System:} Focusing on open-book settings, we utilize the corresponding Wikipedia corpora constructed specifically for the HOVER and FEVEROUS as knowledge sources. To simulate real-world systems, we implement a two-layer retrieval system. The first layer employs BM25 \cite{bm25} as the sparse retrieval algorithm. The second layer combines a Bi-Encoder model (bge-m3) with a Reranker (bge-reranker-v2-m3) \cite{bge-m3}, refining the search results by filtering out irrelevant documents. When interacting with the two functions described in Section \ref{interaction_function}, we set a constraint of a maximum of 15 retrieved documents or a maximum of 6000 tokens, adhering to the model's maximum input length.
\subsection{Main Results}
The overall performance of VeGraph and the baselines are presented in Table \ref{tab:main_results}. The results are organized into two sections. The first section reports the performance of the baseline models as documented in their works, highlighting their diverse configurations, such as variations in the number of examples used for inference, the underlying backbone models and the retrieval systems employed. These models employ varying configurations, including differences in the number of examples used for inference and the retrieval systems implemented. The second section presents the results of our proposed VeGraph model, alongside the reproduced baselines, which are evaluated under identical configurations. From these experiments, we derive several key insights:
\\
\textbf{VeGraph can effectively verify complex claims}: 
VeGraph consistently outperforms most previous models across various test cases. Notably, on the HoVer dataset—where input claims exhibit substantial complexity—VeGraph demonstrates significant improvements, particularly in multi-hop reasoning tasks. Specifically, it achieves a notable 5-point gain in performance on four-hop claims, highlighting its effectiveness in handling complex claim verification. In contrast to the five-run ensemble strategy employed in ProgramFC, VeGraph utilizes an iterative interaction approach, wherein each iteration builds upon the previous one. This step-by-step reasoning mechanism ensures that the output of one iteration serves as the input for the next, rather than merely aggregating multiple independent predictions. Consequently, the final result is derived from a refined, sequential reasoning process. These findings emphasize the crucial role of interactive disambiguation in our approach, underscoring VeGraph's suitability for verifying intricate claims that require advanced reasoning capabilities.
\\
\textbf{Enhanced entity disambiguation leads to gaining in performance}:
Through the integration of interactive graph representations and the agent-based LLM framework, VeGraph achieves substantial performance gains across multiple benchmark datasets. For instance, in the FEVEROUS dataset, VeGraph surpassed baselines by 2 points in the Disambiguation category and 5 points in the Numerical category. However, VeGraph showed slightly lower performance in the \textit{Multi-hop} category of FEVEROUS. This performance drop compared to ProgramFC is attributed to its use of specialized in-context examples tailored specifically to the FEVEROUS dataset \cite{program_fc}. In fact, unlike complex datasets such as Hover, which require multi-hop entity disambiguation, the multi-hop subset of FEVEROUS only necessitates combining evidence from multiple articles without extensive entity resolution \cite{feverous_dataset}. In contrast, VeGraph employs a generalized reasoning pipeline that consistently integrates entity disambiguation across tasks. While this generalization improves adaptability, it introduces a trade-off in performance (e.g., the Multi-hop partition of FEVEROUS) where task-specific optimization might yield better results. 

\subsection{Ablation Study}
To evaluate the contribution of each component in the proposed VeGraph framework, we conducted an ablation study on the HoVer dataset. Specifically, we analyzed the impact of graph representation for disambiguating entity interactions and the role of multi-step reasoning in decision-making within the LLM-agent framework. We begin by removing the interactive graph component, and then gradually increase the maximum number of disambiguation steps $k$ allowed. The results are presented in Table \ref{tab:ablation}.
\begin{table}[!h]
\centering
\begin{adjustbox}{width=\columnwidth}
\begin{tabular}{l|ccc}
\hline\hline
\textbf{Method} & \textbf{2hop} & \textbf{3hop} & \textbf{4hop} \\
\hline
\textbf{VeGraph - w/o Interactive Graph}  & 64.71 & 56.68 & 43.16 \\
\textbf{VeGraph - 0 step}  & 63.09 & 60.85 & 43.57 \\
\textbf{VeGraph - 1 step}  & 69.09 & 62.34 & 54.83 \\
\textbf{VeGraph - 2 steps} & 69.70 & 63.82 & 57.33 \\
\hline
\textbf{VeGraph - 5 steps} & 69.70 & 66.13 & 58.59 \\
\hline\hline
\end{tabular}
\end{adjustbox}
\caption{Ablation studies on the maximum number of disambiguation steps and the effectiveness of graph representation on Hover dataset.}
\label{tab:ablation}
\end{table}
The results demonstrate that removing graph representation severely degrades performance, especially on more complex claims (e.g., 3-hop and 4-hop). This highlights the importance of graph-based reasoning in VeGraph. Additionally, increasing the number of reasoning steps improves performance, indicating that multi-step decision-making is crucial for verifying complex claims.

\subsection{Interpretability and Error Analysis}
Our proposed VeGraph framework not only enhances the performance of claim verification systems but also offers a high degree of interpretability, which is essential for human comprehension and trust. Examples of these generated reasoning traces are provided in Figure 7 of Appendix B. To evaluate the quality of the reasoning processes and the generated graphs, we conducted a human analysis on 50 failed predictions for each partition (2-hop, 3-hop, 4-hop) of the HOVER dataset, focusing on instances where VeGraph incorrectly predicted the claim's veracity. Human annotators categorized the errors into three primary types, corresponding to different stages of the framework: i) \textbf{Graph Representation Errors:} These occur when VeGraph fails to accurately capture the semantic structure of the claim, resulting in flawed graph representations; ii) \textbf{Entity Resolution Errors:} These arise when the system either fails to disambiguate entities or struggles to correctly identify the entities relevant to the claim; iii) \textbf{Subclaim Errors:} These involve incorrect predictions at the level of individual sub-claims leading to erroneous final verdicts.
\begin{table}[!h]
\centering
\begin{adjustbox}{width=0.92\columnwidth}
\begin{tabular}{l|ccc}
\hline\hline
\textbf{Error Types} & \textbf{2hop} & \textbf{3hop} & \textbf{4hop} \\
\hline
\textbf{Graph Representation}  & 29\% & 15\% & 17\% \\
\textbf{Entity Disambiguation}  & 37\% & 53\% & 45\% \\
\textbf{Subclaims Verification} & 34\% & 32\% & 38\% \\
\hline\hline
\end{tabular}
\end{adjustbox}
\caption{Proportions of incorrectly predicted examples across partitions on the HOVER dataset.}
\label{tab:error_analysis}
\end{table}
\\
As shown in Table \ref{tab:error_analysis}, the error distribution varies across the 2-hop, 3-hop, and 4-hop partitions of the HOVER dataset. Despite few-shot in-context learning strategies being employed, the LLM occasionally encounters challenges in constructing accurate graph representations, particularly when processing complex claims. The increasing complexity of multi-hop claims (e.g., 3-hop and 4-hop) further exacerbates issues in entity disambiguation, as a larger number of ambiguous entities complicates the retrieval of relevant documents. Even after multiple interaction cycles, entity disambiguation may remain incomplete, affecting the overall reasoning process. These limitations in both graph construction and entity resolution propagate through the framework, leading to reduced accuracy in the final verdicts, particularly in multi-hop scenarios. Additionally, another source of error comes from failed interactions with the knowledge base, where unresolved triplets mislead the retrieval system, underscoring the critical importance of retrieval performance.

\section{Conclusion}
This study presents VeGraph, a novel claim verification framework using the concept of interactive graph representation incorporating LLM agent technology to identify ambiguous entities in terms of multi-step reasoning of veracity predictions. Specifically, the input claim first is decomposed into a set of triplets. These triplets are then identified with ambiguous entities and verified of fact interactively using the proposed agent LLM pipeline. The experiment on two well-known benchmark claim verification datasets indicates promising results of VeGraph for claim verification tasks, especially in the case of complex claims. 

\section*{Limitations}
While the proposed framework enhances performance in disambiguating entities and verifying sub-claims, it imposes computational overhead due to its frequent reliance on large language models. This increased demand for computational resources can introduce latency, posing challenges for real-world applications that require rapid response times.

Despite their advanced reasoning capabilities, LLMs are prone to errors and may exhibit biases toward certain types of content. This highlights the need for careful deployment, especially in fact-checking systems, where biased or incorrect outputs could lead to misinformation. Developing effective mechanisms to detect, control, and mitigate these biases remains an open challenge for future research.

Another limitation lies in the dataset used for our experiments, which predominantly focuses on explicit reasoning. Although the framework incorporates self-analysis and structured representation, real-world claims often require processing implicit information, adding complexity beyond the current design. Addressing this gap will be a crucial direction for future work, enabling the framework to manage nuanced reasoning better and improve its practical applicability.

\bibliography{custom}

\appendix

\section{Additional Experiments}
\subsection{Cost Analysis}
To provide an understanding of the computational overhead, we conducted a cost analysis on the HoVer dataset. Table~\ref{tab:cost_analysis} summarizes the comparative results of VeGraph and baseline models across metrics, including the number of LLM calls, knowledge base (KB) interactions, and total inference time.

\begin{table}[H]
\centering
\begin{adjustbox}{width=0.95\columnwidth}
\begin{tabular}{lccc}
\hline
\hline
\textbf{Metric}              & \textbf{2-hop} & \textbf{3-hop} & \textbf{4-hop} \\ \hline
\multicolumn{4}{l}{\textbf{VeGraph}}                                 \\
LLM Calls                    & 6.16           & 8.2            & 10.04         \\
KB Interactions              & 3.87           & 4.63           & 5.6           \\
Inference Time (s)           & 9.19           & 10.25          & 12.84         \\ \hline
\multicolumn{4}{l}{\textbf{FOLK}}                                    \\
LLM Calls                    & 4.47           & 4.93           & 5.49          \\
KB Interactions              & 2.47           & 2.93           & 3.49          \\
Inference Time (s)           & 7.98           & 9.35           & 11.09         \\ \hline
\multicolumn{4}{l}{\textbf{ProgramFC}}                               \\
LLM Calls                    & 3.39           & 4.17           & 5.02          \\
KB Interactions              & 2.39           & 3.17           & 4.02          \\
Inference Time (s)           & 6.37           & 7.17           & 8.58          \\ \hline
\hline
\end{tabular}
\end{adjustbox}
\caption{Cost Analysis on HoVer Dataset}
\label{tab:cost_analysis}
\end{table}

As illustrated in Table~\ref{tab:cost_analysis}, VeGraph demonstrates superior reasoning capabilities at a higher computational cost. The disambiguation process, essential for resolving hidden entities and ensuring accurate multi-hop reasoning, contributes significantly to this overhead, primarily due to iterative KB interactions. Specifically, VeGraph's total computational time exceeds that of ProgramFC by approximately 40--50\% and FOLK by 10--15\%. This increase is strongly correlated with the number of reasoning hops, as the frequency of both LLM calls and KB interactions escalates with the query's complexity. While this trade-off reflects the computational demands of VeGraph's advanced reasoning mechanisms, it also underscores the potential for future research to mitigate these costs. Optimizing the disambiguation process and improving overall system efficiency are promising directions to reduce overhead while preserving VeGraph's robust reasoning performance.

\section{Evaluation of Entity Disambiguation Performance}
To evaluate the effectiveness of our method in resolving ambiguous entities, we report the average number of entity resolution requests to the knowledge base (KB) on the HoVer dataset, along with the corresponding success rates of our approach.
\begin{table}[!h]
\centering
\begin{adjustbox}{width=0.95\columnwidth}
\begin{tabular}{lccc}
\hline
\hline
\textbf{Method} & \textbf{2hop} & \textbf{3hop} & \textbf{4hop} \\ \hline
VeGraph   & 1.16 (72\%) & 2.11 (67\%) & 3.08 (70\%) \\
ProgramFC & 0.57        & 1.24        & 1.6         \\ \hline
\hline
\end{tabular}
\end{adjustbox}
\caption{Number of entity resolving requests on HoVer dataset}
\label{tab:entity_resolution_performance}
\end{table}

As shown in Table~\ref{tab:entity_resolution_performance}, approximately 30\% of the requests to the KB failed to resolve the entity. This highlights the importance of the iterative reasoning strategy employed in our VeGraph framework to find the entity. Additionally, the increase in the number of successfully resolved entities demonstrates the enhancement of VeGraph over ProgramFC.

\section{Examples}
\begin{figure*}[!h]
    \centering 
    \includegraphics[width=\textwidth]{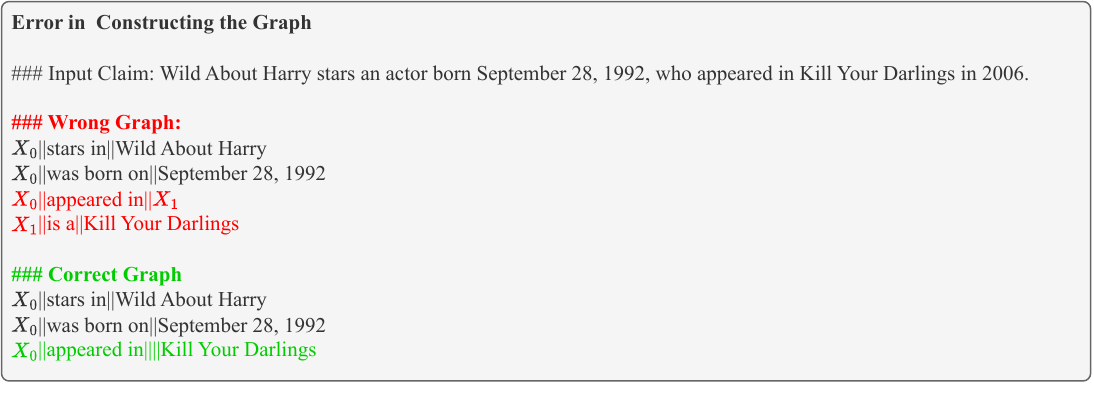} 
    \caption{Incorrect Example 1}
    \label{fig:wrong_example_1}
\end{figure*}

\begin{figure*}[!h]
    \centering 
    \includegraphics[width=\textwidth]{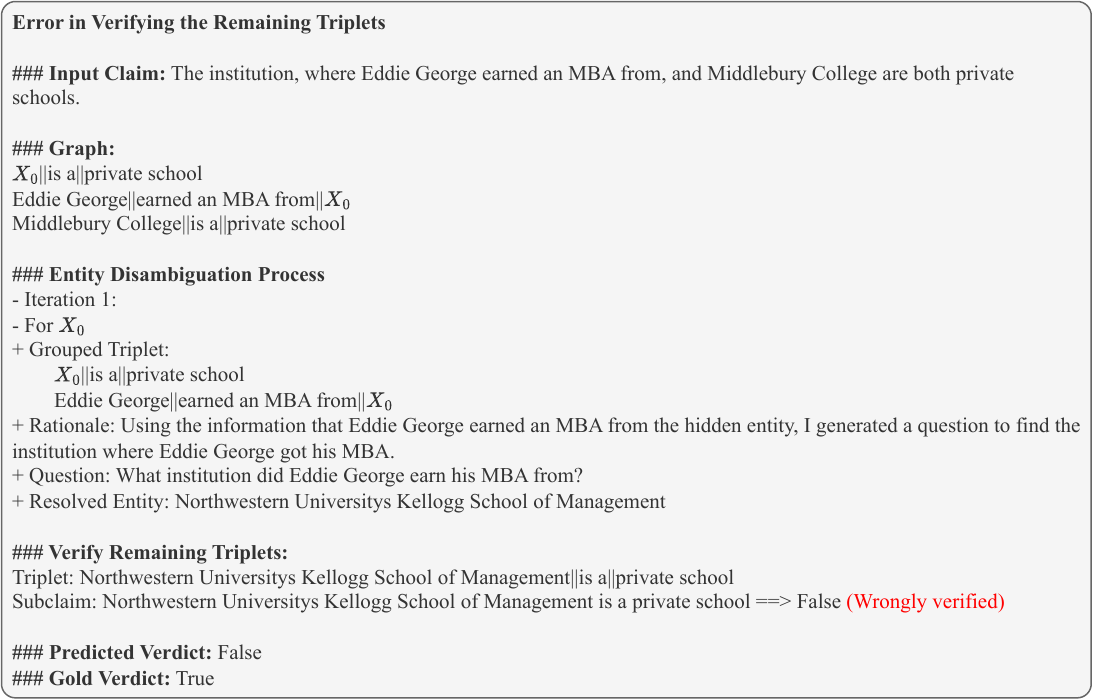} 
    \caption{Incorrect Example 2}
    \label{fig:wrong_example_2}
\end{figure*}

\begin{figure*}[!h]
    \centering 
    \includegraphics[width=\textwidth]{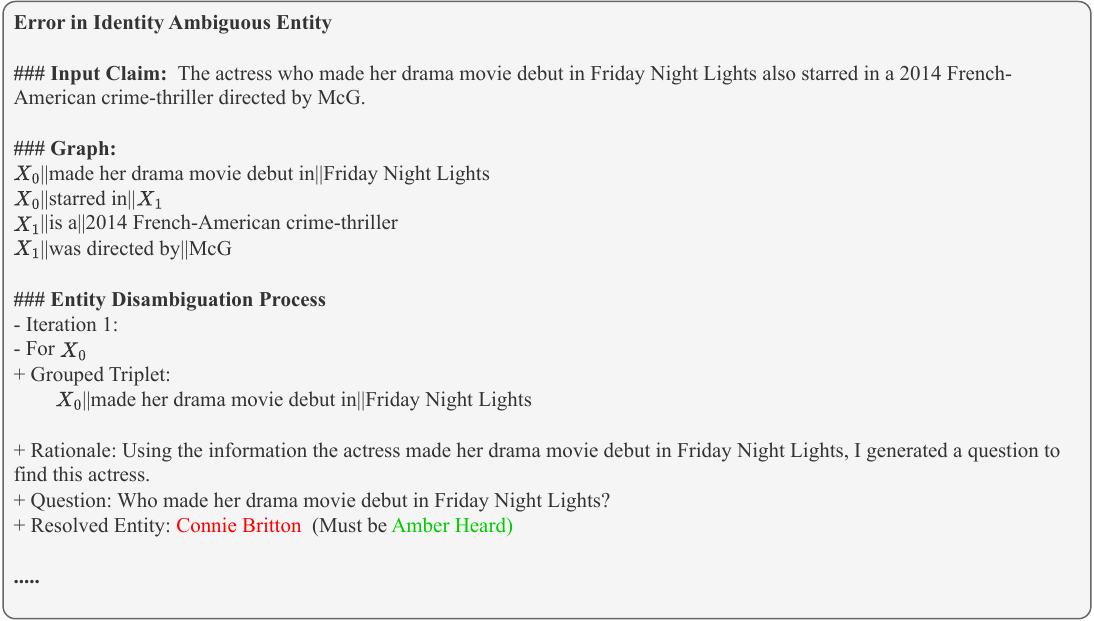} 
    \caption{Incorrect Example 3}
    \label{fig:wrong_example_3}
\end{figure*}

\begin{figure*}[!h]
    \centering 
    \includegraphics[width=\textwidth]{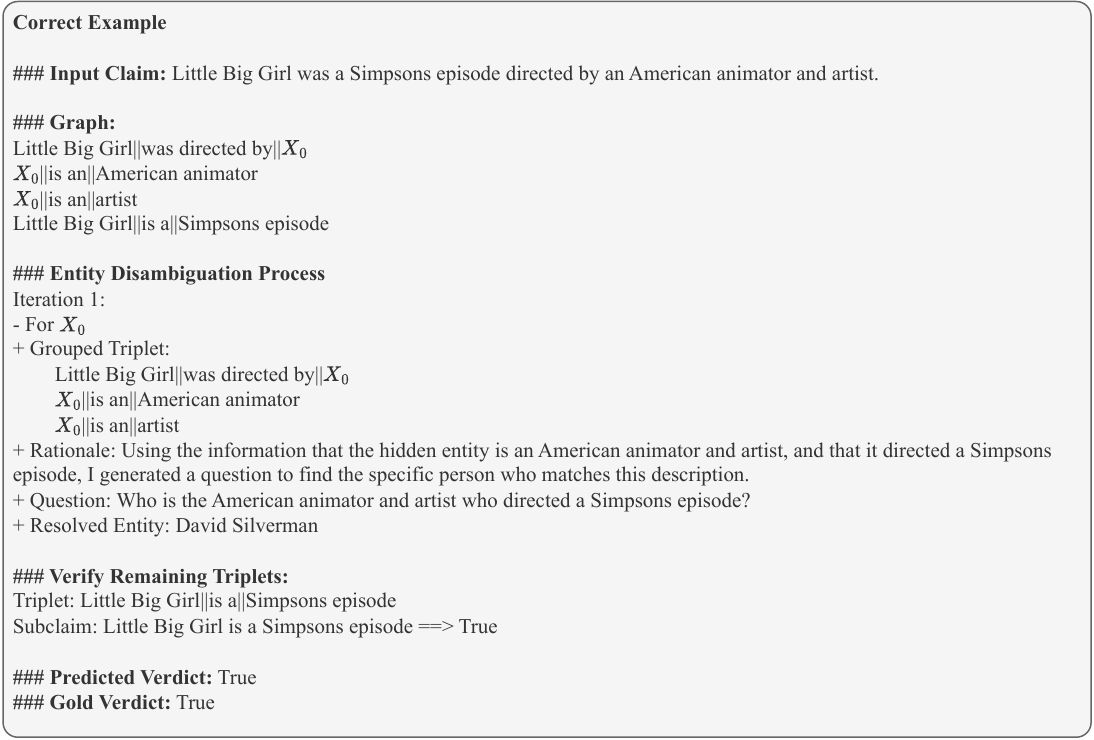} 
    \caption{Correct Example Output 1}
    \label{fig:correct_example_1}
\end{figure*}

\begin{figure*}[!h]
    \centering 
    \includegraphics[width=\textwidth]{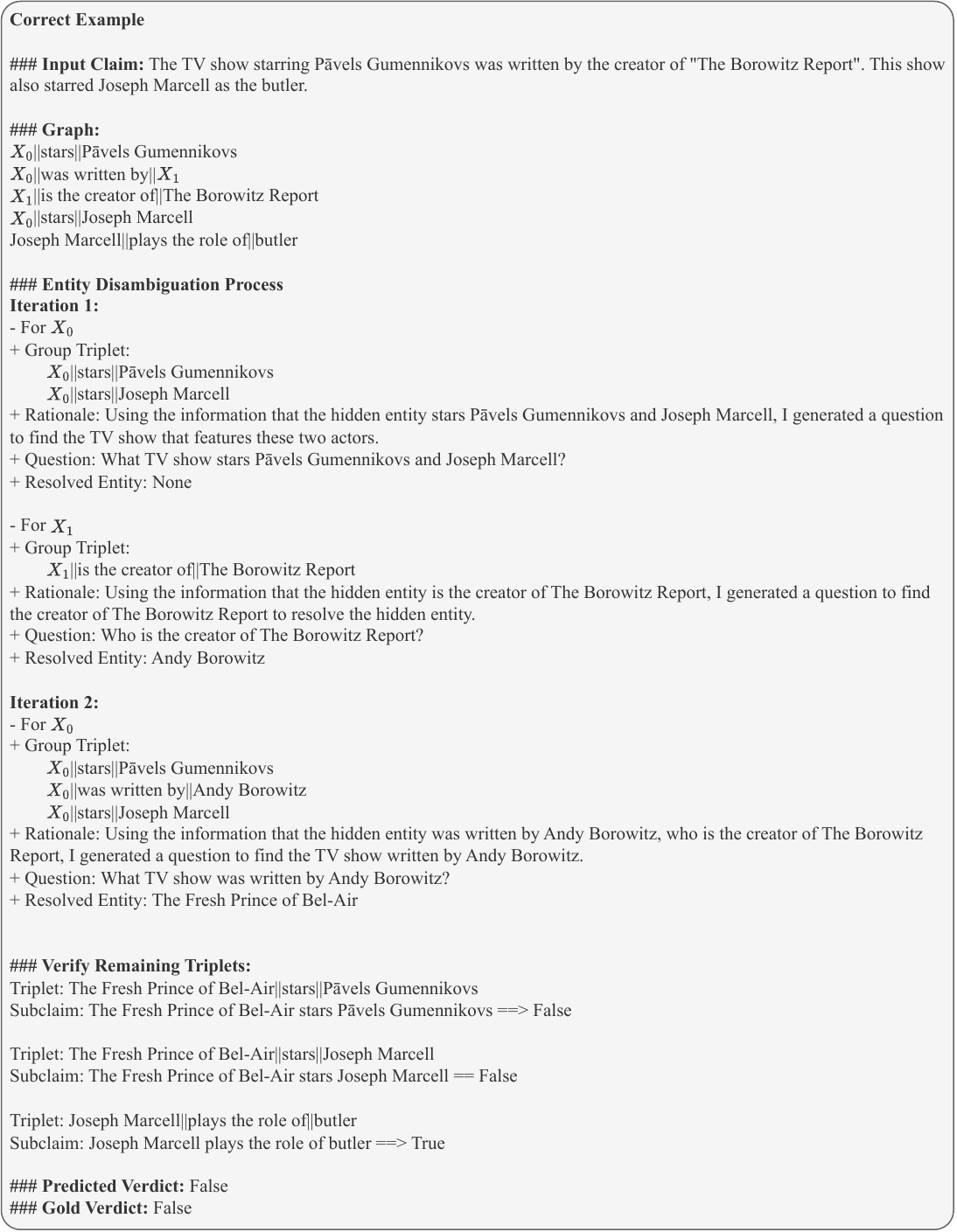} 
    \caption{Correct Example Output 2}
    \label{fig:correct_example_2}
\end{figure*}

We provide illustrative examples to offer a more intuitive understanding of the framework. Figures \ref{fig:wrong_example_1}, \ref{fig:wrong_example_2}, and \ref{fig:wrong_example_3} showcase three distinct error types as discussed in the main section, highlighting common challenges and failure cases. In contrast, Figures \ref{fig:correct_example_1} and \ref{fig:correct_example_2} present correct examples, demonstrating the reasoning traces and outputs at each stage of the framework. These examples collectively serve to clarify the functionality and robustness of the proposed approach.


\section {Prompt Templates}
\label{sec:appendix}
\label{prompt_template}
\begin{figure*}[!h]
    \centering 
    \includegraphics[width=\textwidth]{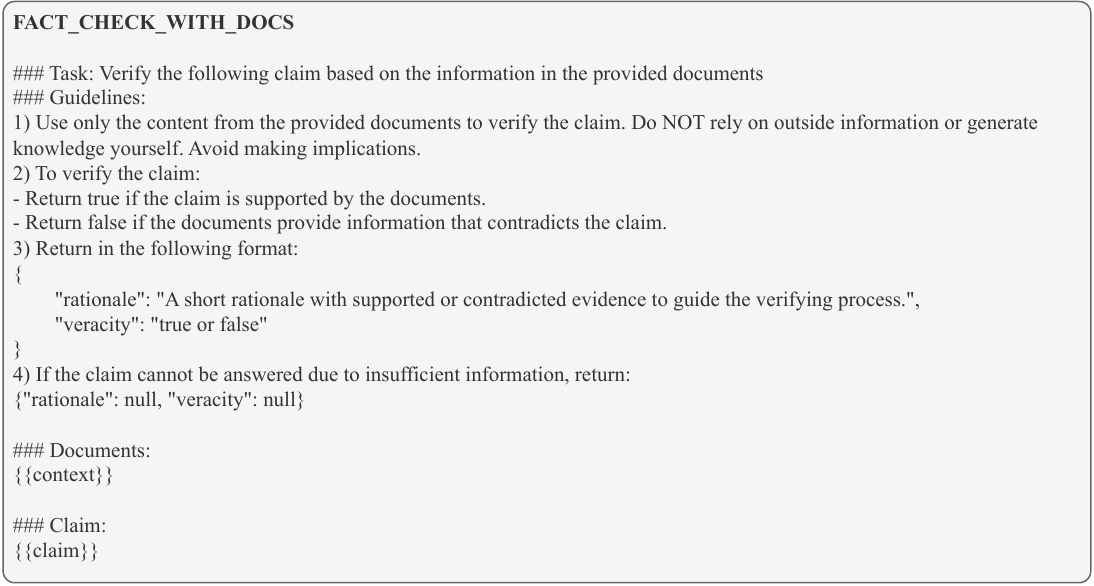} 
    \caption{Prompt template to find related section content from articles.}
    \label{fig:fact_check_with_docs}
\end{figure*}

\begin{figure*}[!h]
    \centering 
    \includegraphics[width=\textwidth]{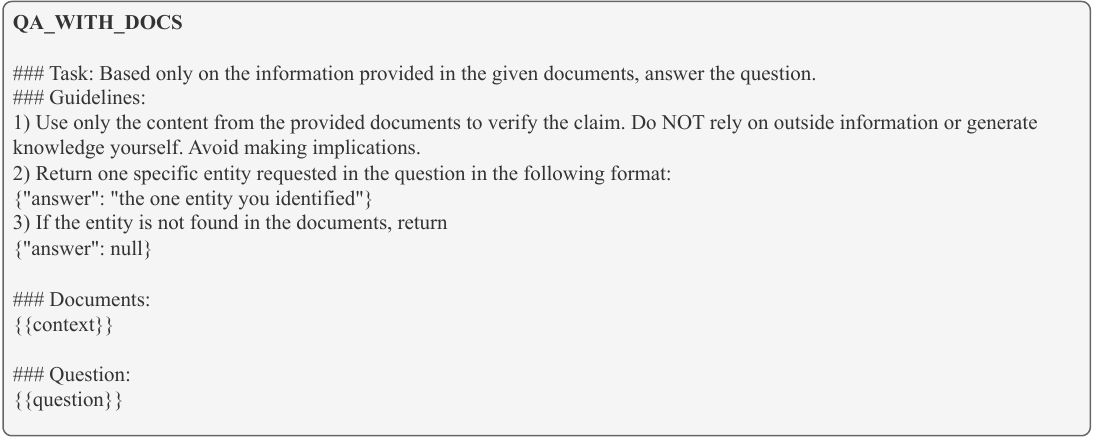} 
    \caption{Prompt template to find related section content from articles.}
    \label{fig:qa_with_docs}
\end{figure*}

\begin{figure*}[!h]
    \centering 
    \includegraphics[width=\textwidth]{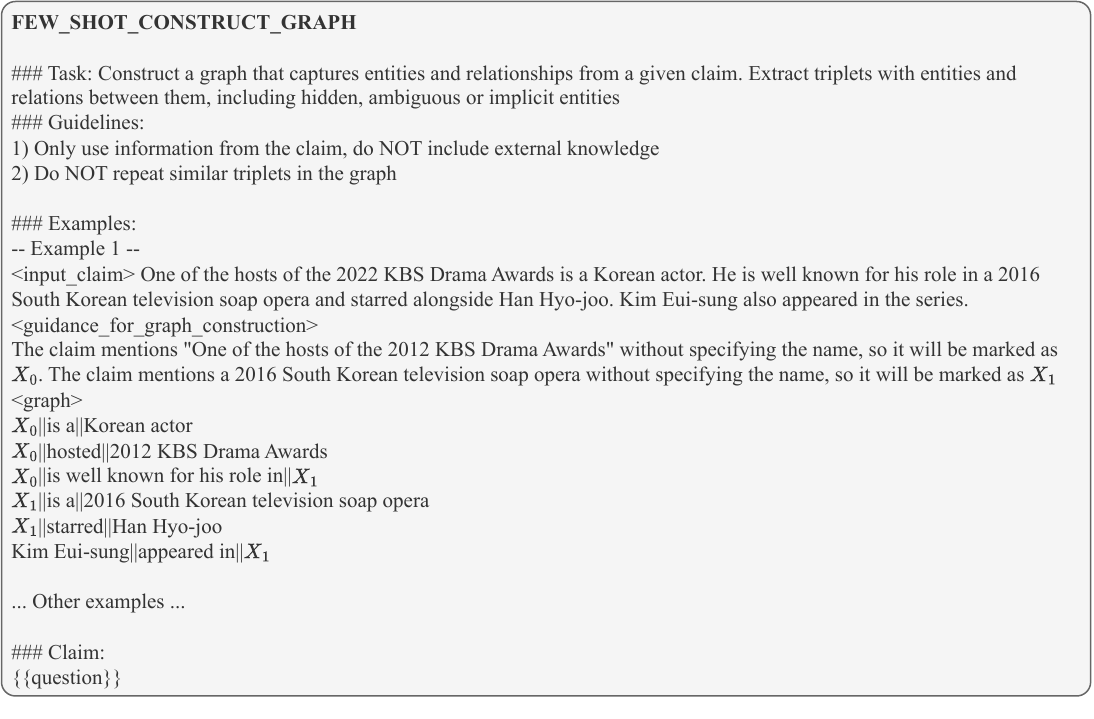} 
    \caption{Prompt template to find related section content from articles.}
    \label{fig:few_shot_construct_graph}
\end{figure*}

\begin{figure*}[!h]
    \centering 
    \includegraphics[width=\textwidth]{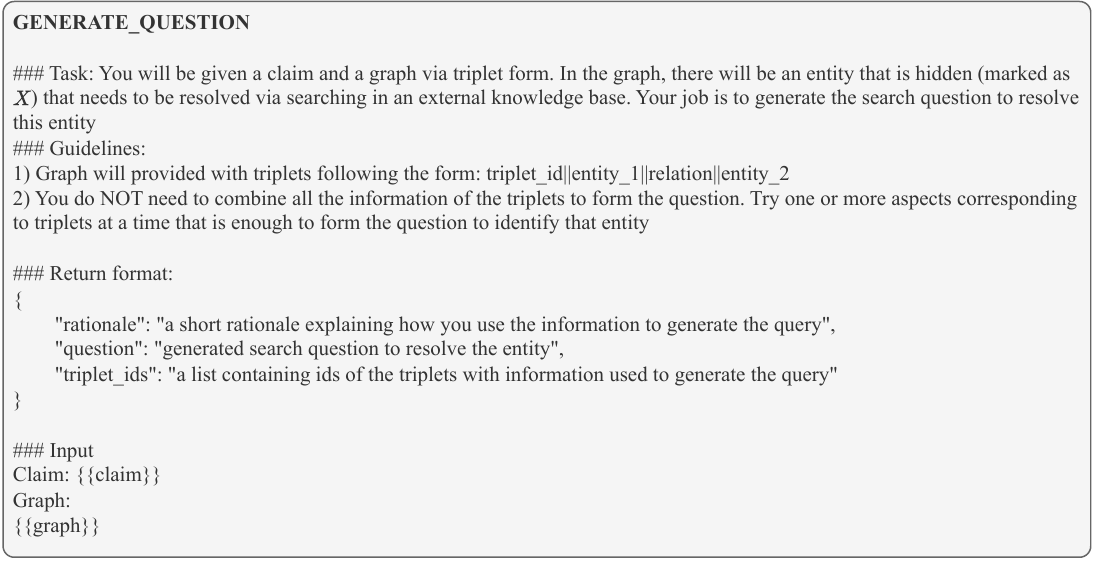} 
    \caption{Prompt template to find related section content from articles.}
    \label{fig:generate_question}
\end{figure*}

\begin{figure*}[!h]
    \centering 
    \includegraphics[width=\textwidth]{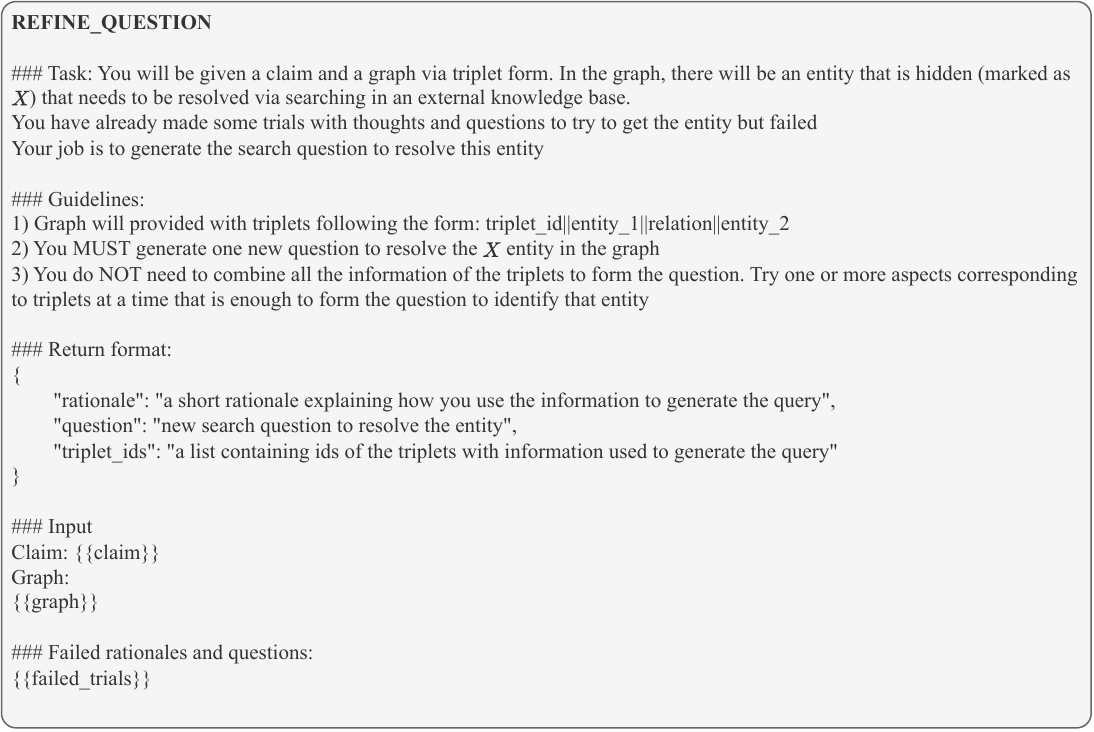} 
    \caption{Prompt template to find related section content from articles.}
    \label{fig:refine_question}
\end{figure*}

\begin{figure*}[!h]
    \centering 
    \includegraphics[width=\textwidth]{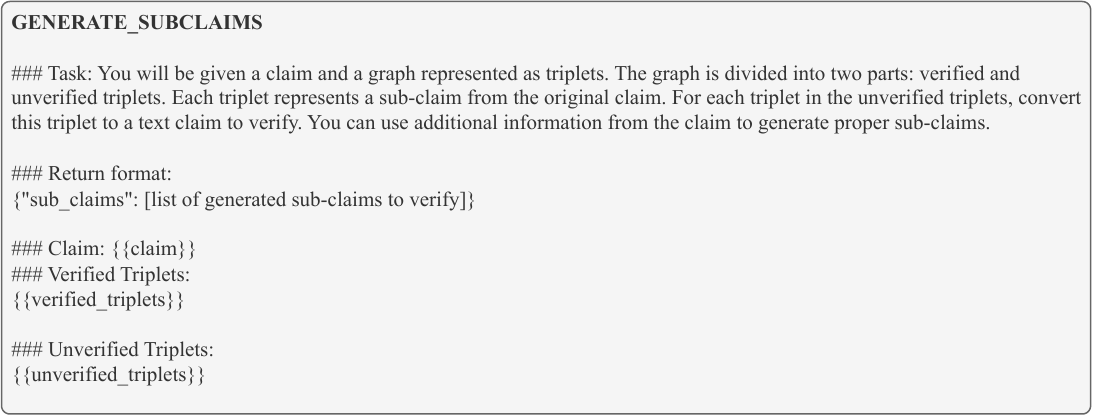} 
    \caption{Prompt template to find related section content from articles.}
    \label{fig:generate_subclaims}
\end{figure*}

For better reproducibility, we present all prompt templates in the appendix. Below is a quick reference list outlining the prompt templates and their usages:
\begin{itemize}
    \item Figure \ref{fig:fact_check_with_docs}: Verify a claim based on the information within a set of documents. 
    \item Figure \ref{fig:qa_with_docs}: Extract an entity within a set of documents that satisfies a question.
    \item Figure \ref{fig:few_shot_construct_graph}: Construct a graph representation of the input claim.
    \item Figure \ref{fig:generate_question}: Generate a question to resolve the ambiguous entity from the given graph triplets and claim.
    \item Figure \ref{fig:refine_question}: Refine failed questions and generate a new question to resolve the ambiguous entity from the given graph triplets and claim.
    \item Figure \ref{fig:generate_subclaims}: Generate sub-claims each corresponding to a graph triplet.
\end{itemize}
All prompts are zero-shot, except for the prompt in Figure \ref{fig:few_shot_construct_graph}, which uses few-shot demonstrations to better guide the LLM to perform the task. These prompts were chosen because they perform effectively in practice.


\end{document}